\documentclass{article}

\usepackage[preprint]{neurips_2025}

\usepackage[utf8]{inputenc} 
\usepackage[T1]{fontenc}    
\usepackage{hyperref}       
\usepackage{url}            
\usepackage{booktabs}       
\usepackage{amsfonts}       
\usepackage{nicefrac}       
\usepackage{microtype}      
\usepackage{xcolor}         
\usepackage{graphicx}       
\usepackage{amsmath}

\title{Evolution imposes an inductive bias that alters and accelerates learning dynamics}

\author{
  Benjamin Midler \\
  Princeton Neuroscience Institute\\
  Princeton, NJ 08540 \\
  \texttt{bm9751@princeton.edu} \\
  \And
  Alejandro Pan Vazquez \\
  Princeton Neuroscience Institute \\
  Princeton, NJ 08540 \\
  \texttt{apv2@princeton.edu}
}

\begin{document}

\maketitle

\begin{abstract}

    The learning dynamics of biological brains and artificial neural networks are of interest to both neuroscience and machine learning. A key difference between them is that neural networks are often trained from a randomly initialized state whereas each brain is the product of generations of evolutionary optimization, yielding innate structures that enable few-shot learning and inbuilt reflexes. Artificial neural networks, by contrast, require non-ethological quantities of training data to attain comparable performance. To investigate the effect of evolutionary optimization on the learning dynamics of neural networks, we combined algorithms simulating natural selection and online learning to produce a method for evolutionarily conditioning artificial neural networks, and applied it to both reinforcement and supervised learning contexts. We found the evolutionary conditioning algorithm, by itself, performs comparably to an unoptimized baseline. However, evolutionarily conditioned networks show signs of unique and latent learning dynamics, and can be rapidly fine-tuned to optimal performance. These results suggest evolution constitutes an inductive bias that tunes neural systems to enable rapid learning.

\end{abstract}

\section{Introduction}

\begin{figure}
      \centering
      \includegraphics[width=0.55\textwidth]{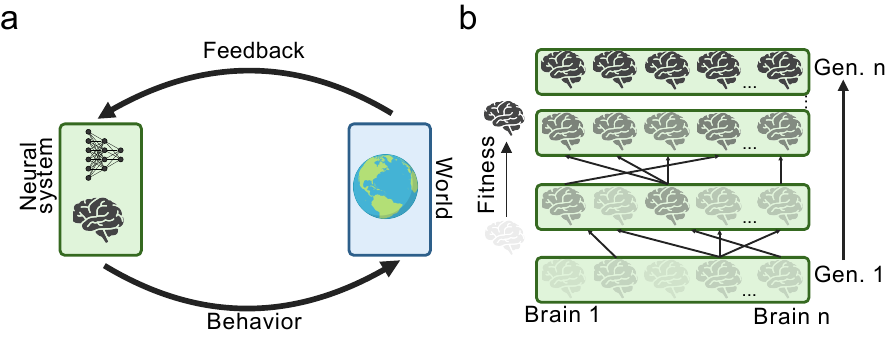}
      \caption{Brains are optimized by natural selection and online learning. (A) Brains and artificial neural networks engage in online learning by producing behaviors and receiving environmental feedback to guide learning. (B) The fitness of biological brains is optimized over evolutionary timescales by natural selection. This constitutes an additional form of optimization on the initial state of a brain, endowing it with inbuilt structures for innate reflexes and enhanced learning potential. In biology, evolutionary optimization and online learning co-exist as evolutionary pressures act upon the genetic information that produces a new brain that then engages in online learning.}
      \label{fig1}
    \end{figure}

    Both neuroscience and machine learning focus on neural systems—biological brains or artificial neural networks—that learn using online mechanisms like synaptic plasticity or backpropagation-based gradient descent \cite{Rumelhart1986-jf, Benfenati2007-tu}. However, neural networks are often trained from a random initial state with no preexisting knowledge, whereas biological brains, through natural selection, have undergone an additional form of optimization that may produce innate neural structures (Fig. \ref{fig1}).

    The objective function of natural selection is to maximize fitness, or the ability to survive and reproduce \cite{Darwin1859-zh}. A cornerstone of the brain’s contribution to fitness is its ability to engage in online learning: a brain that can adapt to its environment will likely outlast one that cannot. As such, there is evolutionary pressure in favor of developing a more plastic neural system that can rapidly adapt to its environment. The effect of online learning on the speed and course of evolution is known as the Baldwin effect \cite{Baldwin1896-ob, Morgan1896-wg}. In computational terms, it can be thought of as reciprocal optimization loops: evolution optimizes the initial state of the network to maximize learning potential, and online learning, in turn, factors into the fitness objective function that guides future evolutionary development.

    There is an important contrast between formulating evolution’s impact on learning as optimizing the initial condition of a network and how it is often used in deep learning, namely through genetic algorithms \cite{Holland1975-wv, Miikkulainen2025-mw, Stanley2019-sw, Yao1999-hm}. These algorithms, which task-optimize neural networks by evaluating a population of networks on an objective function and seeding the next generation by applying stochastic noise to the weights of the best performers, are able to learn complex tasks comparably to backpropagation-based gradient descent \cite{Such2017-wj, Salimans2017-on}. In this application of evolutionary logic to machine learning, natural selection is the sole optimization mechanism and is thus analogous to freezing synaptic plasticity at birth. This is in stark contrast to the biological influence evolution has on the brain, as it acts upon the heritable information used to develop a new organism that may then engage in online learning before passing-on heritable information to the next generation. Crucially, online learning has no direct impact on the heritable information transferred between generations; its influence is strictly limited to the fitness objective function that determines the quantity of offspring produced by an individual.

    It is worth noting that there are many examples of genetically encoded innate neural computations that manifest through, for example, neuronal connectivity patterns, dendritic morphology, and firing dynamics \cite{Meng2021-eg}. The drosophila brain, for instance, is known to be highly stereotyped, with much of its connectivity driven by transcription factor expression \cite{Ozel2022-ik, Yoo2023-bc}. Additionally, even for sophisticated mammalian behaviors, the line between innateness and learning may be blurred, as many social behaviors have both innate and learned components \cite{Wei2021-yq}. As such, we do not claim that evolution cannot nor does not optimize neural systems to perform complex operations—this is clearly false—merely that there is an interplay between these two learning mechanisms that yields unique dynamics.
    
    There have been various efforts to more accurately model brain evolution by, for instance, modeling sequential steps of genetic and gradient descent-based optimization \cite{Xiaodong2018-ox}, introducing a compression step to simulate a genetic bottleneck \cite{Shuvaev2024-xr}, and by iteratively setting and freezing binary connections \cite{Hinton1987-hi}. Each of these efforts demonstrated improvements in optimization speed, consistent with the Baldwin effect; however, they involve the direct transfer of learned weights from one generation to the next. This is a Lamarckian formulation in which acquired traits are transferred between generations. While recent advances in genetics have not ruled Lamarckian evolution impossible \cite{Koonin2009-ag}, it is not a primary mechanism of evolutionary optimization. The relationship between evolutionary optimization and online learning—and the impact this interplay has on a neural network—thus remains unclear.
    
    To address this, we evolutionarily condition artificial neural networks by using a genetic algorithm and gradient descent in distinct steps. Our approach selectively applies evolutionary pressure to information passed between generations, and limits the role of online learning to parameterizing the objective function. This addresses the gap left by prior work as learned network weights are not transferred between generations. We found our model to yield unique latent learning dynamics and significant improvements in the speed of learning, thereby suggesting evolution imposes an inductive bias that promotes the rapid acquisition of new knowledge.

\section{Results}

\subsection{Evolutionary conditioning implementation}

    To assess the effect of merging computational models of evolutionary optimization and gradient descent, we developed a new algorithmic process, dubbed evolutionary conditioning (EC). EC is architecturally agnostic and we applied it to networks in reinforcement and supervised learning contexts.
    
    The core of the EC process is similar to that of existing genetic algorithms: we treat the weights of a deep neural network as a genotype which is inherited by subsequent generations \cite{Miikkulainen2025-mw, Yao1999-hm, Such2017-wj}. The key departure from a standard genetic algorithm is that, rather than directly evaluating each network in the generation, we independently fine-tune network weights on a rollout of the task using backpropagation-based gradient descent. We then evaluate the fine-tuned networks to determine fitness and probabilistically spawn the next generation by applying gaussian noise to the weights of the best performing parent networks. Crucially, we use the parent’s weights prior to fine-tuning to form the child network. This dissociates the evolutionary and gradient-descent optimization loops, thereby constraining the influence online learning has on the evolutionary process to the fitness objective function. In effect, this causes the genetic algorithm to optimize the network’s potential to learn the task rather than directly optimizing the network’s ability to perform the task. While these two objectives can converge—it is easy to learn something if you already possess the knowledge innately—we find this to not be the case.

\subsection{Reinforcement learning reveals latent behavioral structure}

\begin{figure}
      \centering
      \includegraphics[width=0.8\textwidth]{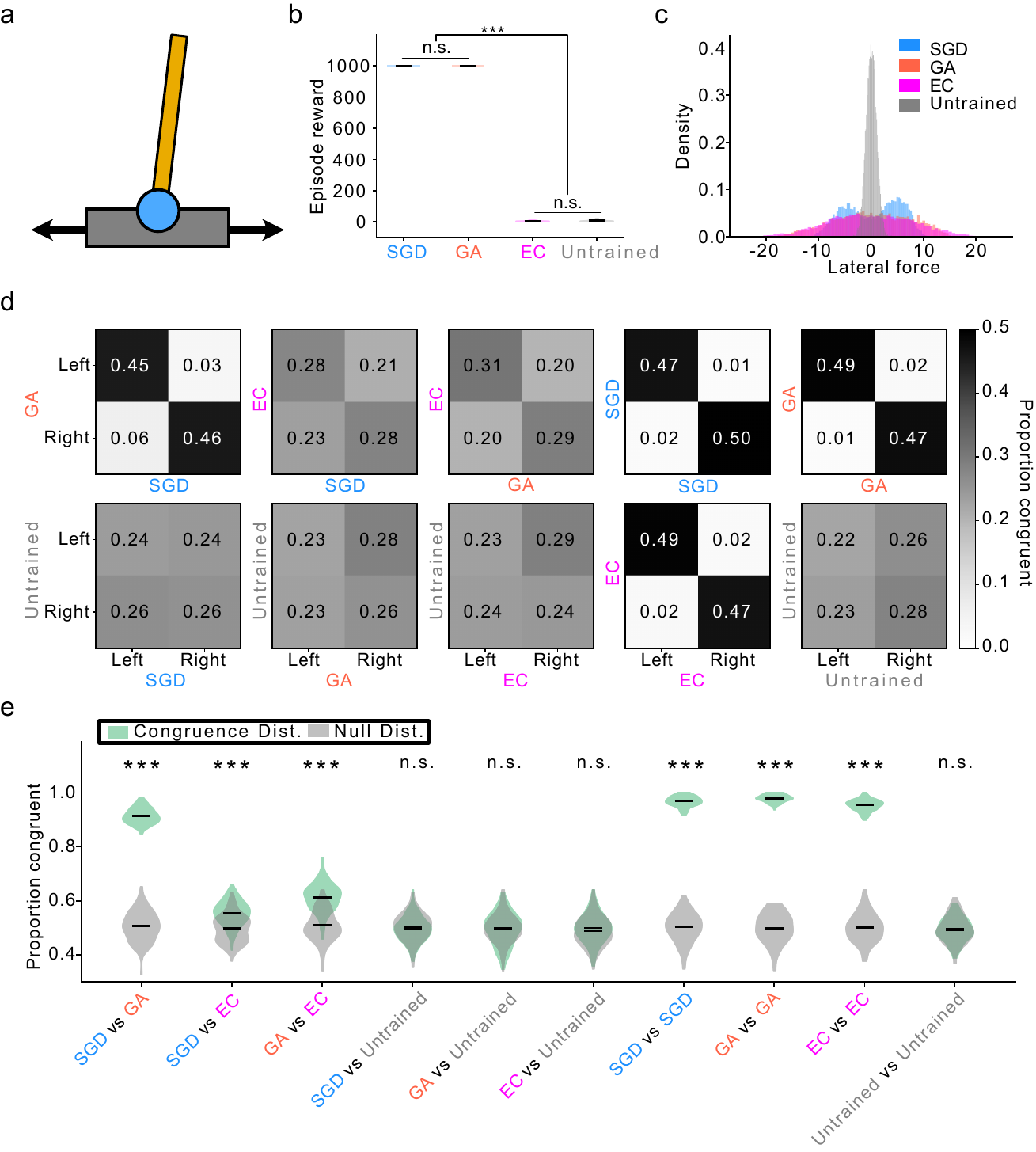}
      \caption{EC fails to task-optimize but shows signs of latent learning. (A) Schematic of the reinforcement learning problem. A pole is balancing on a platform that can be moved laterally. Networks are trained to move the platform to keep the pole balancing upright. (B) Task performance per model. Each reward is a timestep during which the pole is balanced. Max per episode is 1,000, n=50 episodes (Mann-Whitney U test ***p<1e-18 Bonferroni corrected). (C) Distribution of network actions in response to 10,000 uniformly sampled task states (vectors of cart position, velocity, pole angle, and angular velocity). EC behavior is more similar to the GA even though performance is no different than an untrained network. Note: GA and EC distributions are highly overlapping. (D) Example behavioral congruence matrices between each model pair. (E) Congruence distributions (green) versus shuffled null (gray), n=100 matrices as in D. EC is internally consistent and significantly similar to successfully trained networks, indicating latent learning (Mann-Whitney U test ***p<4e-12 Bonferroni corrected).}
      \label{fig2}
    \end{figure}

    We first assessed EC on a reinforcement learning task in which the network must learn to balance a pole by applying lateral force to the platform on which the pole stands \cite{Todorov2012-er, Barto1983-rb}. This is a simple and well-defined benchmark for motor control, which is a fundamental operation for any brain (Fig. \ref{fig2}a).
    
    We trained a neural network to perform this task using both a standard genetic algorithm (GA) and backpropagation-based gradient descent (SGD)—thus decomposing EC into its constituent evolutionary and online learning processes. Both algorithms were able to train the network to perform at ceiling, defined as the pole remaining upright for 1,000 time steps (both models mean=1,000, SD +/- 0 over 50 episodes; Fig. \ref{fig2}b). We then trained the network using EC for 2,000 generations—equal to the number used for the GA—and with a roll-out of 100 episodes for fine-tuning—5\% of the total used for SGD. Whereas both SGD and the GA were able to consistently learn the task, EC consistently failed: its performance was indistinguishable from an untrained network (EC mean=5.28, SD +/-1.47; untrained mean=6.88, SD +/-4.06; Mann-Whitney U test Bonferroni corrected p=1.0 over 50 episodes; Fig. \ref{fig2}b). This prompted us to investigate the networks’ behavior. We uniformly sampled 10,000 task observations—vectors detailing features like cart position and velocity—and recorded each network’s behavioral responses to the observations. In other words: given a random task state, what does the network do? This demonstrated that, even though the EC and untrained networks performed indistinguishably, EC network behavior followed a flatter distribution far more similar to that of the GA network (EC vs untrained network Kolmogorov-Smirnov distance=0.4103; EC vs GA Kolmogorov-Smirnov distance=0.0426; Fig. \ref{fig2}c), and that the GA and SGD networks, despite both performing at ceiling, displayed significantly different behavioral distributions (Kolmogorov-Smirnov distance=0.1254, Bonferroni corrected p<4e-67).
    
    To better understand if these distributional differences are reflected in underlying behavioral strategy, we again passed sampled task states to each network and recorded behavioral responses, but this time we did so simultaneously for two networks to assess their congruence. We abstracted over the magnitude of network output, thus binarizing behavior as a move to the left or the right. Behavioral congruence between networks was expressed as a matrix, with matrix cells giving the proportion of observation inputs that produced a given combination of left or right outputs from the networks (Fig. \ref{fig2}d). The matrix diagonal was then summed to give the proportional congruence between networks. We repeated this 100 times to form congruence distributions for each combination of networks, and produced empirical null distributions by shuffling the observation order for each network combination (Fig. \ref{fig2}e). As expected, SGD and GA networks had both high rates of internal congruence and were significantly congruent—albeit slightly less so—with each other (SGD internal congruence Welch’s t-test Bonferroni corrected p<2e-166; GA internal congruence Welch’s t-test Bonferroni corrected p<7e-156; SGD vs GA congruence Welch’s t-test Bonferroni corrected p<3e-142). Of greater interest was the EC network having a significantly elevated rate of internal congruence and significantly elevated congruence with both SGD and GA networks (EC internal congruence Welch’s t-test Bonferroni corrected p<2e-151; EC vs SGD Welch’s t-test Bonferroni corrected p<4e-12; EC vs GA Welch’s t-test Bonferroni corrected p<2e-32). These results show the EC network, despite performing the task indistinguishably from an untrained baseline, evidenced a consistent behavioral strategy that overlapped significantly with networks fully optimized on the motor control task. This indicates that nesting dissociated loops of evolutionary optimization and online learning yield a unique optimization problem with signs of latent learning.

\subsection{Evolutionary conditioning dynamics are distinct from learning dynamics}

\begin{figure}
      \centering
      \includegraphics[width=0.8\textwidth]{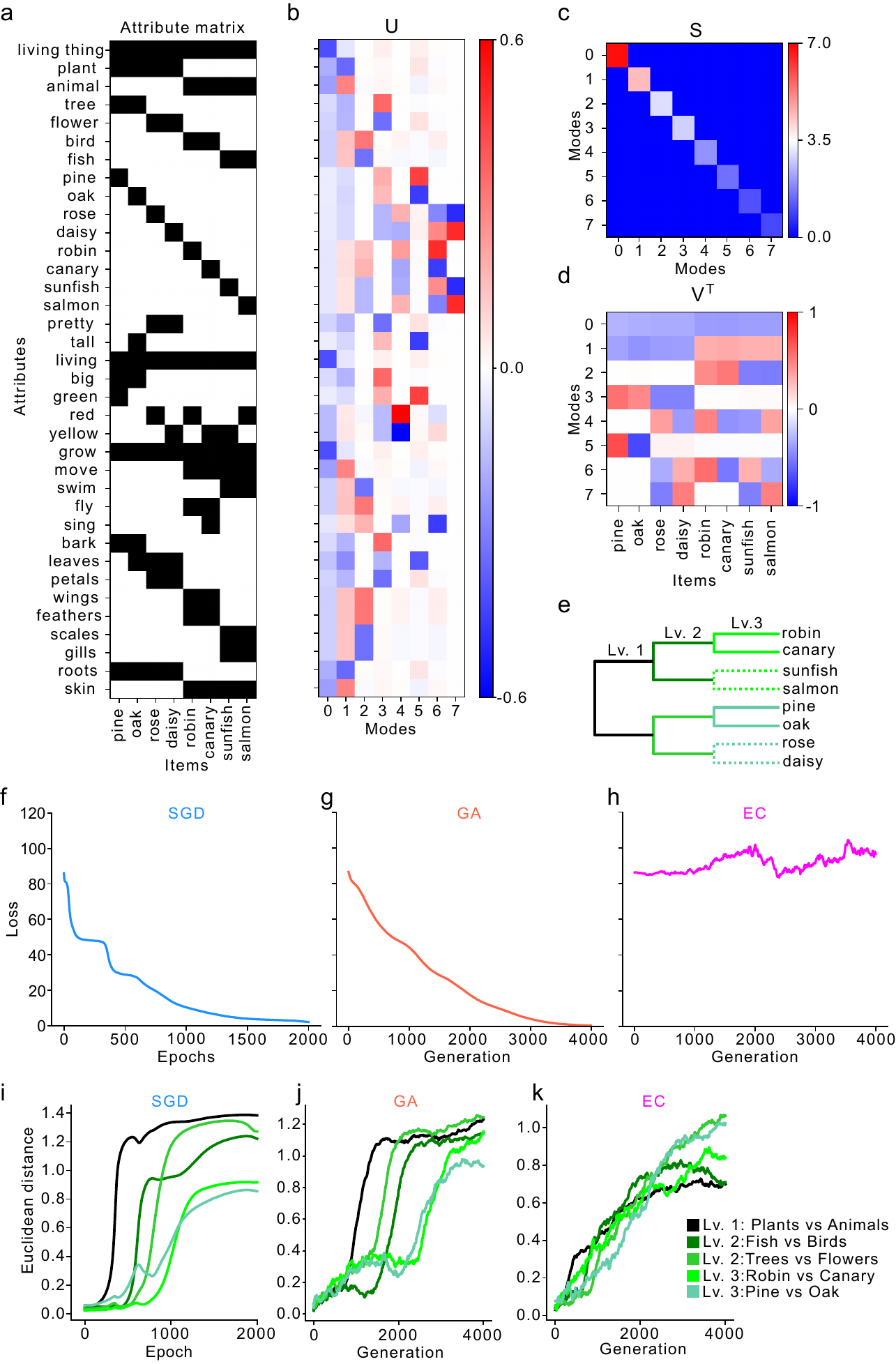}
      \caption{EC representational dynamics are unique and do not respond to known regularities in training data. (A) The items-to-attributes dataset used in the semantic cognition dataset. (B) The U matrix produced by the singular value decomposition of the dataset. The U matrix shows the loadings of each attribute per mode. (C) The S matrix produced by the singular value decomposition of the dataset. The values of the diagonal, or singular values, correspond to the precedence of the modes. (D) The VT matrix produced by the singular value decomposition of the dataset. It shows the loadings of each item onto the modes, illustrating how the modes correspond to categorical differentiations (e.g. mode 1 splits plants and animals, mode 2 splits birds and fish). (E) Schematic of dataset similarity structure divided into categorical levels. Solid branches are those shown in I-K, dotted branches are not shown but are equivalent to those from the same hierarchical level. (F-H) Loss curves of each model during training. Both SGD and GA task-optimize, but not EC. (I-K) Representational dynamics of each model throughout training. Both SGD and GA show sequential learning dynamics in which superordinate categories are learned before subordinate. EC, however, learns to distinguish all categorical levels concurrently.}
      \label{fig3}
    \end{figure}

    Reinforcement learning is a fundamental paradigm in neuroscience and machine learning as it is central to how animals and artificial neural networks learn behavioral responses to environmental challenges \cite{Sutton1998-rm}. Interpreting network dynamics during reinforcement learning, however, is difficult \cite{Dethise2019-fg}. Therefore, in order to clarify how EC network dynamics may be distinct from those of SGD and GA networks, we applied all three to a sequence-to-sequence supervised learning problem in which a network learns associations between a set of items and their attributes (Fig. \ref{fig3}a). This is a semantic cognition task with several properties useful for forming mechanistic hypotheses for how a neural network acquires information \cite{Saxe2019-ub, McClelland2003-at}. Specifically, we applied singular value decomposition to the items-to-attributes dataset, yielding a set of modes, the weightings of each item onto the modes, and the singular values that correspond to mode precedence (Fig. \ref{fig3}b-d). These modes correspond to the hierarchical similarity structure between the items in the dataset, arranging them into superordinate (e.g. plants and animals) and subordinate (e.g. birds and fish) categories (Fig. \ref{fig3}e). Networks trained on this semantic cognition task acquire information sequentially, following mode order: networks first learn to differentiate the highest-order categories before progressing to lower-orders and finally differentiating the items themselves \cite{Saxe2019-ub, McClelland2003-at}.

    We trained networks on the semantic cognition task using SGD and the GA, confirming that the algorithms could task-optimize (Fig. \ref{fig3}f-g). Learning dynamics were assessed by taking the vectors of hidden-layer activity evoked by categorical inputs and computing the euclidean distance between them. When a level of the categorical hierarchy is learned, the euclidean distance between the category vectors increases. Using this method, we confirmed that the learning dynamics of networks trained with SGD and the GA followed the sequential order given by the dataset’s similarity structure and singular value decomposition (Fig. \ref{fig3}i-j). We then trained the network using EC. As with the reinforcement learning problem, the EC-trained network did not task-optimize: over the course of 4,000 generations its loss remained flat or increased (Fig. \ref{fig3}h). Moreover, the EC network’s learning dynamics were markedly different from those of the GA and SGD networks. Whereas those algorithms yielded the expected sequence of learning higher-order categories before lower-orders, EC concurrently differentiated all levels of the hierarchy (Fig. \ref{fig3}k). This indicates EC not only produces latent learning dynamics, which was also indicated by the reinforcement learning task, but that those latent dynamics are distinct from other learning algorithms and do not respond to the same mathematical regularities in the training data.

\subsection{Evolutionary conditioning facilitates rapid learning}

    Thus far we have found that EC displays latent learning dynamics in reinforcement and supervised learning tasks. A core prediction of the Baldwin effect is that the evolutionary process yields more rapid online learning \cite{Dennett1993-oz, Hinton1987-hi}. To test this prediction, we took EC networks spaced 250 generations apart and used backpropagation-based gradient descent to fine-tune them on the semantic cognition task (Fig. \ref{fig4}a). This is analogous to an organism being born the product of generations of evolutionary optimization and then engaging in online learning. We found that the number of fine-tuning epochs required for EC networks to reach criterion decreased monotonically from 1,597 (SD +/-159) for a network with no EC—only fine-tuning—to 148 (SD +/-0) for a network with 4,000 generations of EC (Mann-Whitney U test p<5e-38; Fig. \ref{fig4}b-e). This order of magnitude increase in training speed constitutes computational support for the Baldwin effect, and indicates evolutionary optimization can accelerate online learning even when no gradient information nor gradient-optimized weights are transferred between generations.

\subsection{Evolutionary conditioning alters gradient descent learning dynamics}

\begin{figure}
      \centering
      \includegraphics[width=0.8\textwidth]{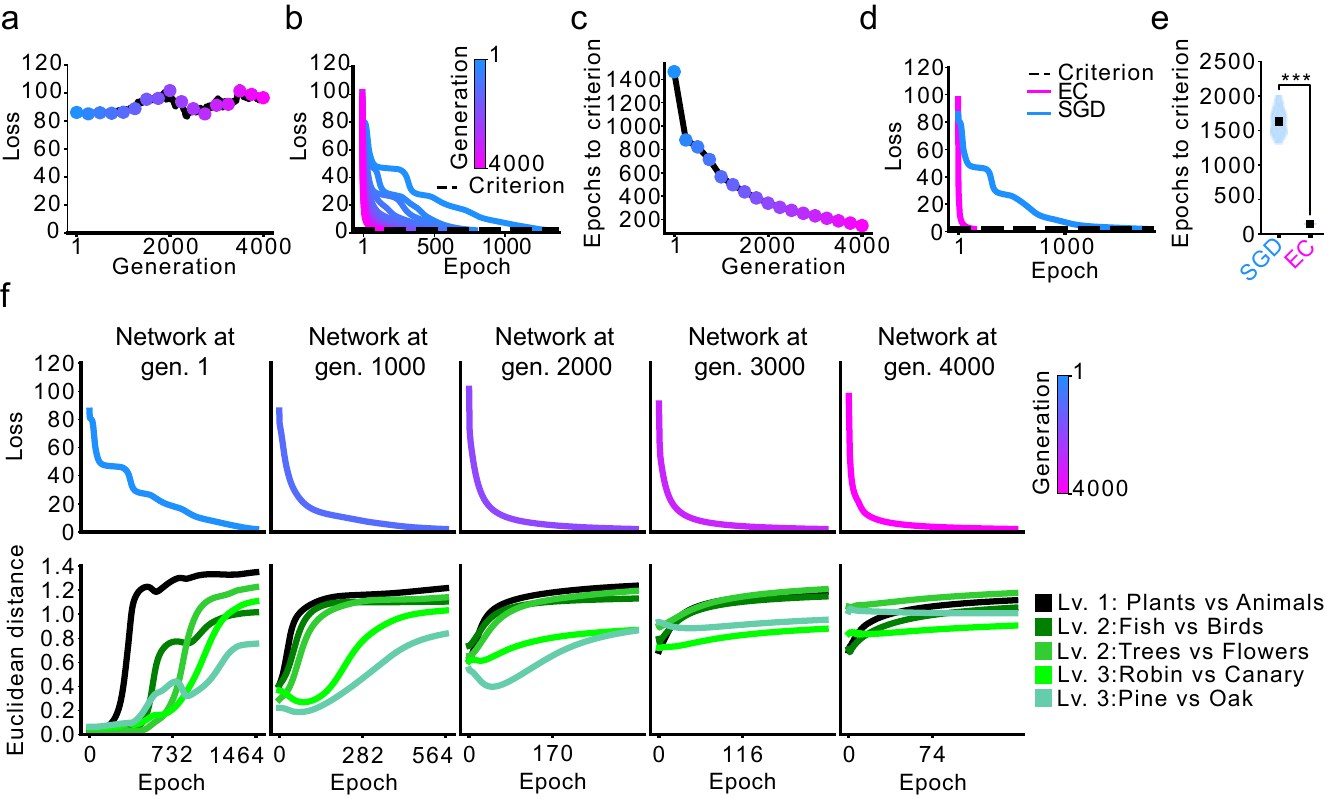}
      \caption{EC speeds fine-tuning and alters gradient descent representational dynamics. (A) EC loss curve with colored dots indicating network snapshots taken form throughout training. (B) Loss curves of network snapshots from A being trained to criterion. (C) Same as in B but each network snapshot is expressed as epochs of training required to reach criterion. Even though loss during EC is flat or elevated (shown in A), it accelerates fine-tuning by an order of magnitude. (D) Loss curves for SGD and EC networks trained to criterion. (E) Comparing epochs to train to criterion for a network trained via SGD and a network pre-trained with EC then fine-tuned (Mann-Whitney U test Bonferroni corrected ***p<5e-38). (F) Loss curves (top) and representational dynamics (bottom) for network snapshots from throughout EC as they were fine-tuned. Dynamics are unique not only during EC, but during fine-tuning as well.}
      \label{fig4}
    \end{figure}

    Though the EC process itself displays unique representational dynamics distinct from those of networks trained with SGD or the GA, we were curious how EC, as a form of pre-training, altered the learning dynamics of gradient descent during fine-tuning. Does EC push the network to a parameter subspace that, when fine-tuned, shows the same learning dynamics as the SGD or GA networks but at a faster timescale, or do the dynamics remain unique? Thus, as a final test, we compared the learning dynamics of networks from throughout the EC process as they were fine-tuned using gradient descent (Fig. \ref{fig4}f). Doing so revealed that, early in EC, learning dynamics followed a sequential pattern similar to the SGD and GA networks (Fig. \ref{fig4}f left column). As EC progressed, however, the pattern changed. At intermediate generations the top two levels of the hierarchy were collapsed into a single learning phase (Fig. \ref{fig4}f middle columns), and, at late generations, representational distances for all categorical levels remained stable and intermixed (Fig. \ref{fig4}f right column). This is despite the fact that, as loss remains elevated throughout EC, the networks saw similar levels of performance improvement during fine-tuning. These simulation results further support the notion that evolution introduces latent representational dynamics into the weights of neural networks, and offers a potential explanation for why brains optimized by generations of evolution can display distinct and more rapid learning dynamics than computational models.

\section{Discussion}

    Evolutionary models have yielded useful insights into the developmental course of neurobiology and cognition \cite{Miikkulainen2025-mw, Stanley2019-sw}. For instance, genetic algorithms have been used to form de novo hypotheses on the emergence of the command neurons found in \textit{aplysia}, lobsters, and crayfish, demonstrating that computational models of evolution can contain biological insight \cite{Miikkulainen2025-mw, Aharonov-Barki2001-di}. As a complementary point, a recent study ported a connectome of \textit{C. elegans} into a simulated worm, finding that the biological synaptic weights constitute a high-performing and data-efficient baseline for training the model worm to swim, demonstrating biology can be a useful prior in machine learning \cite{Bhattasali2022-tm}.
    
    This latter example is indicative of the \textit{tabula rasa} problem: that artificial neural networks—which purport to be at least superficially based on the brain—are often randomly initialized as a blank slate \cite{Zador2019-uc}. This is in contrast to the generations of evolutionary optimization that endow biological brains with an innate structure upon which online learning builds. We used a novel model of neural development that dissociates evolutionary optimization from online learning to investigate the evolution-driven emergence of innate structure in artificial neural networks. We demonstrate that evolution conditions a neural network to display latent behavioral and computational dynamics that, while undetectable in baseline performance, constitute an inductive bias that significantly alters learning dynamics, violates theories of how networks respond to mathematical regularities in training data, and speeds learning by an order of magnitude. Evolutionary conditioning may therefore help explain how biological brains are able to learn new information from only a few examples whereas computational models are notoriously data inefficient \cite{Kaplan2020-kk, Hafner2019-fi, Graves2016-hn}.
    
    While we apply this model to both reinforcement and supervised learning tasks, our focus was on using small, relatively tractable problems that would provide mechanistic insight into computational and behavioral dynamics. Going forward, it will be useful to investigate this formulation of neural evolution by applying it to a broad range of computational problems and models, to use evolutionary algorithms that jointly act on network weights and architectures \cite{Stanley2002-rs}, and to concurrently evolve the physiology of an agent and its neural network controller \cite{Gupta2021-cn}. Doing so will help clarify the level of neural structure upon which evolutionary pressures operate. In our model, online learning and evolutionary optimization were applied to the same weights—albeit in distinct optimization steps. While network nodes in the semantic cognition simulation ought to be thought of as coarse connections between abstract cognitive modules rather than synapses between neurons—an important distinction as synaptic weights are not genetically encoded but coarse axon wiring can be \cite{Kerstjens2022-xf, Hassan2015-pu}—it will be important to understand how evolutionary optimization and online learning can cooperate across multiple levels of brain structure to produce optimal solutions.
    
    In sum, we propose a novel algorithm that models evolution and online learning as distinct steps. We demonstrate how this algorithm conditions artificial neural networks to display latent behavioral and computational dynamics that do not follow known patterns of network learning, yet accelerates task-optimization by an order of magnitude. These results suggest evolution imposes an inductive bias on neural networks that significantly alters and accelerates learning dynamics.


\begin{ack}

    This work was not supported by any dedicated funding. Princeton University has a patent pending for technology related to this manuscript.

\end{ack}



\newpage
\bibliographystyle{plain}
\bibliography{EvolInductiveBias_NeurIPS2025}





\newpage
\appendix

\section{Technical Appendices and Supplementary Material}

\subsection{Reinforcement learning: task}

    The reinforcement learning task was the MuJoCo implementation of the cart pole motor control problem \cite{Towers2024-bp, Barto1983-rb}. We used default environment parameters (uniform sampling of initial state, max episode length=1,000 time steps, reset noise scale=0.01). The task objective is to keep a pole balanced upright by applying lateral force to the platform on which the pole is balancing. A successful time step is defined as |angle|<0.2 radians, and returns a reward of +1. The episode terminates if angle exceeds 0.2 radians, or the maximum episode length is reached, whichever occurs first. The task produces four observation values each time step: cart position along the linear slider (meters), cart velocity (meters/second), pole angle (radians), and pole angular velocity (radians/second). All observation values are continuous and range from negative to positive infinity. The task action space has one value: lateral force applied to the cart (newtons). Force values are continuous between -3 (maximum leftward force) and +3 (maximum rightward force).

\subsection{Reinforcement learning: gradient descent implementation}

    The stochastic gradient descent (SGD) model trained on the cart pole reinforcement learning problem was a deep neural network that takes task observations as an input and produces as an output the mean () and standard deviation () of a distribution that is sampled to produce the model action. Loss is computed by the REINFORCE algorithm which is then used to update network weights via backpropagation \cite{Williams1992-gt, Rumelhart1986-jf}.

    The model was implemented using PyTorch \cite{Paszke2019-be}, and is structured as such:

    \begingroup
        \centering
            \vspace{10pt}
            
            $
            \text{Input: } s \in \mathbb{R}^{\text{obs\_dim}}
            $
    
            $
            \text{Hidden Layer 1: } h_1 = \text{ReLU}(W_1 s+ b_1)
            $
    
            $
            \text{Hidden Layer 2: } h_2 = \text{ReLU}(W_2h_1 + b_2)
            $
    
            $
            \text{Output Layer (Mean): } \mu = W_{\mu} h_2 +b_{\mu}
            $
    
            $
            \text{Output Layer (Log SD): } \log \sigma = \text{Learnable Parameter}
            $
    
            $
            \text{Standard Deviation: } \sigma = \exp(\log\sigma)
            $

            \vspace{10pt}
    \endgroup
    
    The process for producing a model action, \textit{a}, given an input state vector, \textit{s}, and using the neural network as the policy is thus:

    \begingroup
        \centering
            \vspace{10pt}

            $
            \mu, \sigma = \text{policy}(s)
            $

            $
            \text{Distribution: } \mathcal{N}(\mu, \sigma)
            $

            $
            \text{Sampled Action: } a \sim \mathcal{N}(\mu, \sigma)
            $

            $
            \text{Log Probability: } \log p(a) = \sum_{i} \log \mathcal{N}(\mu_i, \sigma_i)
            $

            \vspace{10pt}
    \endgroup

    Note that the log probability of the selected action is computed for subsequent use in the loss function.

    The weights of the network are trained via backpropagation \cite{Rumelhart1986-jf}. To produce the loss via the REINFORCE algorithm, the log probabilities of rewards are summed with return values, which are the discounted rewards received after an action. Specifically:

    \begingroup
        \centering
            \vspace{10pt}

            $
            \text{Returns: } R_t = \sum_{k=0}^{T-t-1} \gamma^i r_{t+i}
            $
            
            $
            \text{Normalized Returns: } \hat{R} = \frac{R - mean(R)}{SD(R) + \epsilon}
            $
            
            $
            \text{Loss: } L = -\sum_{t} \log p(a_t) \hat{R}_t
            $

            \vspace{10pt}
    \endgroup

    Here, $R_t$ is the return value at time step \textit{t}, \textit{T} is the number of total time steps in the episode, $\epsilon$ is a small value to prevent division by zero, and $r_{t+i}$ is reward at time step $t+i$. Models were trained with the discount factor $\gamma=0.99$ and a learning rate of $1e-3$ using the Adam optimizer \cite{Kingma2014-ge}.

\subsection{Reinforcement learning: genetic algorithm implementation}

    The genetic algorithm (GA) model was implemented using the same network architecture as the SGD model: the deep neural network takes task observations as inputs and produces the mean and standard deviation of a distribution that is sampled to produce the action output. The only difference between the models was the training mechanism.

    Specifically, for the GA model, a population of networks was initialized with random weights, constituting the first generation. Each model in the generation was then evaluated on an episode of the cart pole task, with network fitness quantified as the reward received in the episode.

    The reward of each network (policy), $\pi$, is given by the sum of rewards received on each time step, $t$, of the episode, with $T$ being the number of time steps in the episode.

    \begingroup
        \centering
            \vspace{10pt}

            $
            \text{Reward of Policy } \pi_i: R(\pi_i) = \sum_{t=0}^{T} r_t
            $

            \vspace{10pt}
    \endgroup

    To spawn the next generation of networks (child networks), the best performing parent networks are sampled probabilistically—and with replacement—according to their task performance. A selected parent model has gaussian noise applied to its weights to produce a child model.

    \begingroup
        \centering
            \vspace{10pt}

            $
            \text{Reward of Parent Model } \pi_i: R(\pi_i)
            $
            
            $
            \text{Weight of Parent Model } \pi_i: w_i = R(\pi_i) - R_{min} + \epsilon
            $
            
            $
            \text{Probability of Selecting Parent Model } \pi_i: p_i = \frac{w_i}{\sum_{j=1}^{k} w_j}
            $

            \vspace{10pt}
    \endgroup

    Here, $R(\pi_i)$ is the reward for parent model $\pi_i$, $R_{min}$ is the minimum reward received by the $k=3$ top performing parent models, $w_i$ is the probabilistic weight of selecting parent model $\pi_i$, $\epsilon$ is a small constant used to ensure non-zero probability of selecting a parent model for seeding a child model, and $p_i$ is the probability of selecting parent model $\pi_i$ for seeding a child model.

    The process of seeding a child network, $\pi_i'$, from parent network $\pi_i$ consists of adding gaussian noise to the weights of the parent model, with $mean=0$ and $standard deviation=0.01$.

    \begingroup
        \centering
            \vspace{10pt}

            $\pi_i' = \pi_i + \mathcal{N}(0, 0.01)$

            \vspace{10pt}
    \endgroup

    An additional important detail is that we employed generational annealing to create a dynamic schedule of population sizes across generations. This enables large populations at early generations when robustly sampling the space of possible solutions is important, and small populations at later generations for computational efficiency. Specifically, the number of networks at a given generation $g$, $P(g)$, is given by

    \begingroup
        \centering
            \vspace{10pt}

            $\text{Population Size at Generation } g: P(g) = P_{start} \cdot \exp\left(\left(\frac{g}{G-1}\right)^{\lambda} \cdot \log\left(\frac{P_{end}}{P_{start}}\right)\right)$

            \vspace{10pt}
    \endgroup

    Here, $P_{start}$ and $P_{end}$ set the initial and final population sizes and $\lambda$ is a decay term used to set the annealing schedule. It was set to 0.1 whereas $P_{start}$ and $P_{end}$ were 1,000 and 10, respectively. $G$ is the number of generations.

\subsection{Reinforcement learning: evolutionary conditioning implementation}

The evolutionary conditioning (EC) implementation combines those of the SGD and GA models. Put simply, the EC model uses the GA selection mechanism, but, after seeding a new generation, uses the SGD REINFORCE implementation to fine-tune network weights on a rollout of the cart pole task. Networks with fine-tuned weights are then evaluated, as in the GA implementation, and evaluation scores are used to sample networks for seeding the next generation. However, the gaussian noise is applied to network parameters prior to fine-tuning with gradient descent.

\subsection{Reinforcement learning: analyses and statistics}

\subsubsection{Model performance comparison}

    Model performance was compared by initializing 50 random instances of the cart pole task and recording the rewards received by models on the 50 episodes. The same episode initializations were passed to each model in the comparison. Reward per episode is defined as the number of time steps during which $|angle|<0.2$ radians, with a maximum of 1,000 time steps. Two-sided Mann-Whitney U tests were used for pair-wise comparisons between model performances, and p values were Bonferroni corrected for multiple hypotheses by multiplying p values by the number of tests being performed.

\subsubsection{Sampling observations}

    To sample observations for assessing model behavior, 10,000 random numbers were sampled for each of the cart pole task’s four observational variables (cart position, cart velocity, pole angle, and angular velocity). The task technically allows for observational variables to be infinite. To make sampling tractable, we used a range of -1 to 1, which was empirically observed to capture the range of task observations produced by trained and untrained models alike.

\subsubsection{Action distributions}

    To form action distributions produced by the sampled observations, the same observations were passed to each model, and the output action was recorded. Density distributions were plotted as histograms with 75 bins and compared using two-sample Kolmogorov-Smirnov tests. Bonferroni correction was used to compensate for multiple tests.

\subsubsection{Congruency matrices}

    To form action congruency matrices between models, 1,000 observations were sampled as before and passed to two models in concert. Model actions in response to the given observation were recorded as leftward (action < 0) or rightward (action > 0). The cells of the matrix are populated by taking the number of observations that produced each combination of leftward and rightward outputs from the models (e.g. if an observation yielded a rightward move from one model and a leftward move from the other) and were normalized by the number of observations to produce a proportion.

\subsubsection{Congruency distributions}

    To form congruency distributions, the same process as above was repeated 100 times, with each run containing 100 observations. The matrix diagonal for each model pair on each run was then summed to produce the proportion of observations that yielded congruent actions from both models. The distributions are formed of these congruency values over the 100 runs. To make null distributions for statistical comparisons, the process was repeated but with the observations passed to one of the two models randomly shuffled. Statistical significance was established using Welch’s t-test, and p values were corrected for multiple hypotheses testing using Bonferroni correction.

\subsection{Semantic cognition: task}

    The semantic cognition task has been previously described in detail \cite{Saxe2019-ub, McClelland2003-at}. In short, it requires a network to learn associations between items and their attributes, joined by a series of prepositions. The dataset is thus composed of binary input vectors for a set of items and a preposition that links the item to some of its attributes. The preposition input constrains the item-attribute association. For instance, if the input item is “canary” and the input preposition is “can”, the target output attributes are “grow”, “move”, “fly”, and “sing”. The “canary” item has other attributes, such as having wings, but because “can” is the wrong preposition to link “canary” and “wings”, the “wings” output unit is trained to not activate when “canary” and “can” are the inputs (“wings” is an active attribute when “canary” is paired with the preposition “has”). When comparing hidden layer representations between items, comparisons are averaged across all prepositions for the given item.

\subsection{Semantic cognition: stochastic gradient descent implementation}

    The SGD model is a deep neural network that takes items and prepositions as inputs and is trained to output a binary vector for the associated attributes. The model was implemented in PyTorch and used sigmoid nonlinearities, the stochastic gradient descent optimizer, and mean-squared error \cite{Paszke2019-be}. Weights were initialized between $-0.1$ and $+0.1$ and the model was trained with backpropagation using a learning rate of 0.1 \cite{Rumelhart1986-jf}. The model was trained for 2,000 epochs.

\subsection{Semantic cognition: genetic algorithm implementation}

    The GA model used the same architecture, initialization procedure, nonlinearities, and task formulation as the SGD model, only it was trained using the genetic algorithm instead of backpropagation-based gradient descent. For the GA, populations of 100 networks were initialized, evaluated on the semantic condition dataset, with task loss used to represent network fitness that sets sampling probabilities for seeding child networks. This process is the same as it was for the reinforcement learning problem. As with the reinforcement learning formulation, the top $k$ models used to seed the next generation was set to 3 and the standard deviation for the gaussian noise applied to networks was 0.01. The model was trained for 4,000 generations.

\subsection{Semantic cognition: evolutionary conditioning implementation}

    As with the reinforcement learning implementation, the EC model combined the training methodologies of the SGD and GA models such that the GA, instead of using loss of the networks initialized as part of each generation, used the loss of the networks after being fine-tuned on the task with backpropagation-based gradient descent. In this case, networks were fine-tuned on the task for 100 epochs per generation for 4,000 generations. Crucially, the genetic algorithm hyper-parameters (e.g. $k$) were the same as for the GA, and the fine-tuning hyper-parameters (e.g. learning rate) were the same as for the SGD model, with the exception of the number of training epochs.

\subsection{Semantic cognition: analyses and statistics}

\subsubsection{Learning dynamics}

    To assess learning dynamics, vectors of hidden layer activity were extracted for each item input to the model and euclidean distances were computed between vectors to quantify the model’s ability to learn distinctions between input patterns. For categorical comparisons (e.g. plants vs animals), distances were averaged across the constituents of the category. This process was repeated for model snapshots taken throughout training to trace the learning dynamics of the model.

\subsubsection{Training to criterion}

    When assessing the time it takes to train a model to criterion, a network, or a snapshot taken from during training, was fine-tuned using backpropagation-based gradient descent until loss equals a set threshold. For all tests the threshold was set to 2. This value was chosen as it corresponds to accuracy >99\% and SGD models were reliably able to attain it within 2,000 epochs of training and the GA model within 4,000 generations. Testing was performed with slightly different criterion values and the same trends were observed. Statistical significance was assessed using a two-sided Mann-Whitney U test. No correction for multiple hypothesis testing was applied as only one statistical comparison was performed per analysis.

\subsection{Compute resources}

    All analyses and simulations were performed using a consumer-grade laptop (Apple MacBook Pro, 2023) with the M3 Max integrated CPU/GPU and with 48 GB of memory.

    
\end{document}